\documentclass{article}
\usepackage{spconf,amsmath,graphicx}
\usepackage{enumitem}
\usepackage[table,xcdraw]{xcolor}
\usepackage{graphicx}
\usepackage{subcaption}
\usepackage[normalem]{ulem}
\usepackage{caption}
\usepackage{subcaption}
\usepackage{appendix}
\usepackage{multirow}
\useunder{\uline}{\ul}{}

\def\BibTeX{{\rm B\kern-.05em{\sc i\kern-.025em b}\kern-.08em
    T\kern-.1667em\lower.7ex\hbox{E}\kern-.125emX}}

\title{Enhancing Text Annotation through Rationale-Driven Collaborative Few-Shot Prompting}

\name{Jianfei Wu$^{1,4}$, Xubin Wang$^{2,1,3}$, Weijia Jia$^{1,3}$\sthanks{Corresponding author.},~\textit{Fellow,~IEEE}}
\address{$^1$Institute of Artificial Intelligence and Future Networks, Beijing Normal University, Zhuhai, China\\
$^2$ Department of Computer Science, Hong Kong Baptist University, Hong Kong, China\\
$^3$ BNU-HKBU United International College, Zhuhai, China\\
$^4$ College of Artificial Intelligence, Beijing Normal University, Beijing, China
}

\begin{document}
%
\maketitle
\begin{abstract}
The traditional data annotation process is often labor-intensive, time-consuming, and susceptible to human bias, which complicates the management of increasingly complex datasets. This study explores the potential of large language models (LLMs) as automated data annotators to improve efficiency and consistency in annotation tasks. By employing rationale-driven collaborative few-shot prompting techniques, we aim to improve the performance of LLMs in text annotation. We conduct a rigorous evaluation of six LLMs across four benchmark datasets, comparing seven distinct methodologies. Our results demonstrate that collaborative methods consistently outperform traditional few-shot techniques and other baseline approaches, particularly in complex annotation tasks. Our work provides valuable insights and a robust framework for leveraging collaborative learning methods to tackle challenging text annotation tasks.
\end{abstract}

\section{Introduction}

\begin{figure*}[t]
  \centering
  \subcaptionbox{Universal Self-consistency Method\label{fig:image1}}
  {\includegraphics[width=0.41\textwidth]{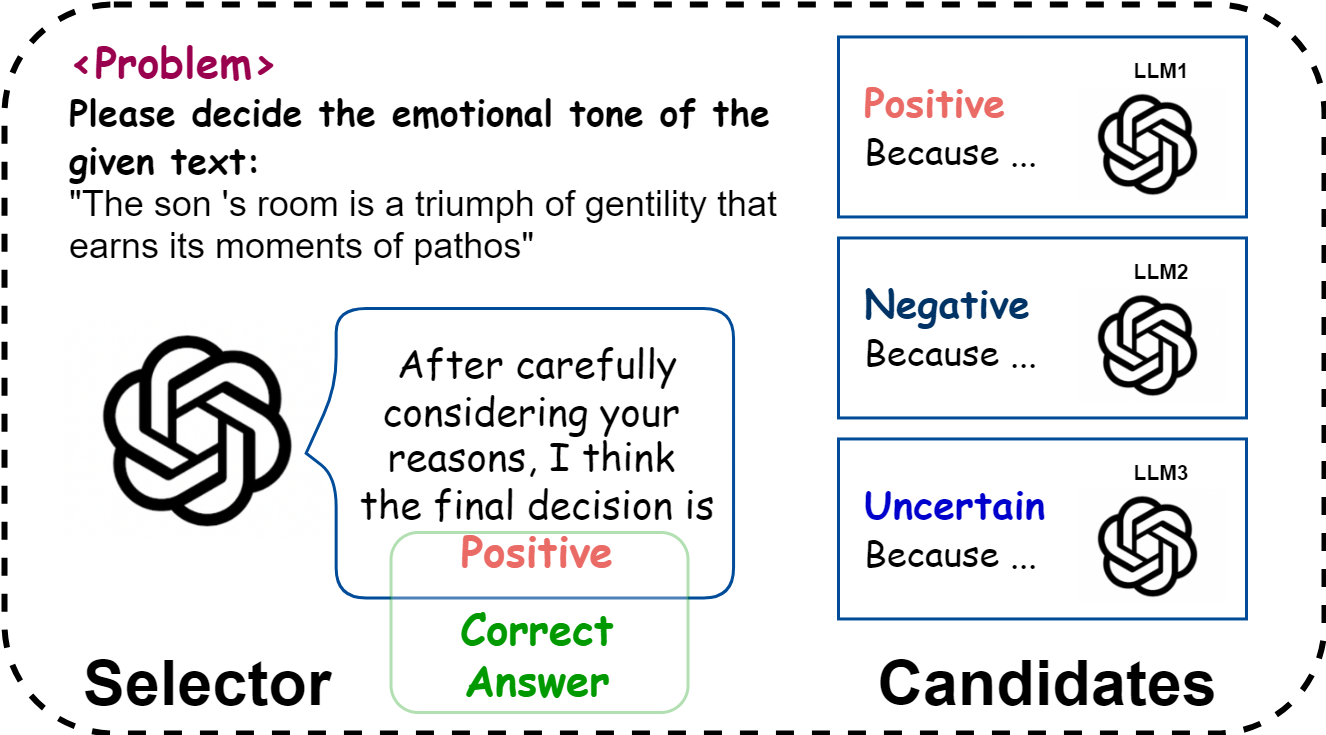}}
  \hfill
  \subcaptionbox{Rationale-driven Collaborative Annotation\label{fig:image2}}
  {\includegraphics[width=0.576\textwidth]{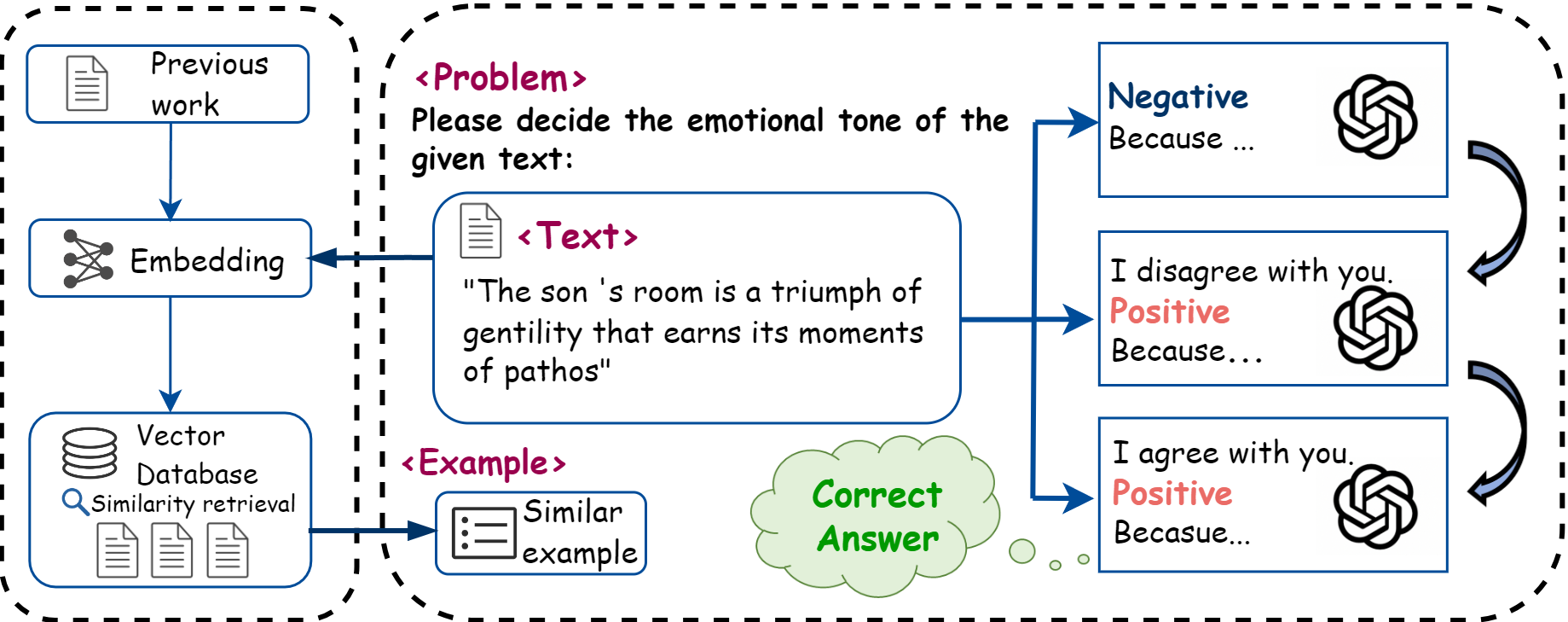}}
  \caption{This figure highlights the differences between universal self-consistency method and the rationale-driven collaborative annotation method. In universal self-consistency, multiple models generate outputs independently, which can result in inconsistencies. In contrast, the rationale-driven collaborative method allows LLMs to perform reasoning consecutively, with each round building on the previous output. }
  \label{fig:two-images}
\end{figure*}

\label{sec:intro}
Data annotation is a critical step in training machine learning models, where the quality and quantity of labeled data directly influence model performance \cite{Survey}. However, traditional annotation methods are plagued by several challenges, including the time-consuming and labor-intensive nature of manual labeling, the potential for human bias and subjectivity, scalability issues, and the difficulties in maintaining annotation quality \cite{autoannotation}. As datasets continue to grow in size and complexity, the demand for efficient and reliable annotation processes has never been greater. 

In this context, LLMs have emerged as a promising solution to address these challenges. By harnessing the capabilities of LLMs, organizations can automate significant portions of the annotation process, drastically reducing the time and effort required to label data. This automation not only enhances efficiency but also improves consistency across annotations, mitigating the variability often associated with human annotators \cite{Outperform, chatgptannotation, raliability}. Moreover, LLMs offer the ability to scale annotation efforts without a corresponding increase in human resources, making it feasible to manage vast amounts of data. They can also be trained to recognize and reduce biases, leading to higher-quality labeled datasets \cite{tan2024large}.  

However, leveraging LLMs for annotation tasks presents notable challenges. The inherently stochastic nature of LLM-generated content, coupled with the phenomenon of hallucination \cite{hallucina}, can compromise their accuracy in performing annotation tasks. While some studies suggest that widely used LLMs, such as ChatGPT, can achieve or even exceed the quality of human annotations \cite{Outperform, biomedicaltask}, many researchers argue that relying solely on LLMs is insufficient for high-quality text annotation, necessitating additional verification processes \cite{autoannotation, icassp1}. 

Inspired by human collective deliberation for resolving ambiguity, we propose a rationale-driven collaborative (RDC) text annotation method using prompt engineering. Unlike traditional methods that rely on the independent outputs of a single model, our approach involves multiple LLMs sequentially annotating a text. Each subsequent LLM receives the annotations from its predecessor and integrates this information to refine its inference process, resulting in more accurate annotations. This collaborative method significantly reduces computational and storage overhead compared to conventional collaborative modeling approaches \cite{brainstorming,self-consistency}. Experimental results indicate that our method outperforms baselines in terms of annotation accuracy. Specifically, the main contributions of our work are as follows:
\begin{itemize}[left=0pt]
\item We propose a rationale-driven collaborative text annotation method based on prompts. This approach leverages intermediate information from the reasoning process more effectively than previous prompting and traditional collaborative methods, thereby improving text annotation quality.

\item We demonstrate that the selection of examples is critical for LLM annotation; matching similar examples significantly enhances annotation effectiveness. Our method reduces the costs associated with manual text annotation while achieving high accuracy, offering a novel direction for automated text annotation research.

\item We find that LLMs can implicitly integrate intermediate processes in rationale-driven collaborative text annotation tasks without the need to store redundant historical dialogues. This approach mitigates the risk of occupying additional generation space, which could otherwise impair model reasoning capabilities.
\end{itemize}

\section{Related Work}
\label{sec:format}

\subsection{Text Annotation via LLMs}
Traditional text annotation processes predominantly rely on manual labor, which is often time-consuming, labor-intensive, and prone to inconsistencies in data quality \cite{autoannotation}. Recent studies suggest that widely utilized LLMs, such as ChatGPT, can outperform human annotators across various annotation tasks, demonstrating a higher degree of consistency compared to manual annotations \cite{Outperform}. Current research on leveraging LLMs for text annotation can be categorized into two primary approaches: the first approach focuses on model adaptation, employing techniques such as fine-tuning to customize models for specific tasks \cite{dataannotationfinetune, Chiang2023CanLL}; the second approach emphasizes prompt engineering, which involves the modification of templates and prompts, as well as the utilization of collaborative strategies across multiple models to enhance annotation accuracy \cite{self-consistency,pdfchat,icassp2}. 

\subsection{Prompts Enhancement in LLMs}
Given the multifaceted capabilities exhibited by LLMs, researchers have made significant strides in transforming generic LLMs into specialized agents capable of executing specific tasks through prompt engineering. Various prompting techniques, including zero-shot prompting \cite{Wei2021FinetunedLM}, few-shot prompting \cite{fewshotlearners, Chen2022ImprovingIF}, Chain-of-Thought (CoT) reasoning \cite{CoT}, and collaborative methodologies \cite{self-consistency,chen2023agentverse}, have been employed to enable LLMs to align more closely with human intentions, systematically analyze problems, and complete reasoning processes. Furthermore, prompts facilitate the integration of reference knowledge beyond basic instructions, allowing LLMs to gather essential clues for generating questions and providing accurate answers. Our work also incorporates several of these prompting techniques to enhance the interpretability and quality of the outputs generated by LLMs. 
 
\section{Methodology}
\label{sec:pagestyle}
The proposed methodology is designed to enhance the quality and efficiency of the annotation process through a structured multi-round collaborative annotation framework. This section outlines the key components of the methodology, including the reasoning process of the LLM, the integration of previous annotations, and the utilization of similar example matching.

\subsection{LLM Reasoning Process}
The initial step in our methodology involves the LLM's reasoning process when presented with a query $ Q $. The LLM generates an answer $ A $ along with a supporting rationale $ R $. This can be mathematically represented as:
\begin{equation}
    (A, R) = \text{LLM}(Q)
\end{equation}

This foundational step is critical, as it establishes the basis for subsequent rounds of annotation. Previous research has often neglected the rationale $ R $, focusing primarily on optimizing the directive to achieve annotation results. This oversight frequently leads to suboptimal quality in the annotations produced.

\subsection{Rationale-Driven Collaborative Annotation Framework}
In our rationale-driven collaborative annotation framework, we build upon the initial response by integrating both the statements awaiting annotation $ S $ and the results from the previous round $ A_{i-1} $ along with their corresponding rationales $ R_{i-1} $. The prompt for the $ i $-th round is defined as:
\begin{equation}
    P_i = (S, A_{i-1}, R_{i-1})
\end{equation}

This integration serves as a reference for the LLM's subsequent annotation, facilitating a collaborative approach that leverages prior outputs to enhance the quality of the current round's annotations.

\subsection{Output Restriction and Error Mitigation}
To improve annotation efficiency and minimize the influence of potential errors from prior rounds, we restrict our reference to the output from the most recent collaborative annotation. The LLM's output for the $ i $-th round is thus defined as:
\begin{equation}
    (A_i, R_i) = \text{LLM}(P_i)
\end{equation}

This approach ensures that the LLM operates without manual intervention or the need for additional prompting techniques, such as CoT prompting, thereby significantly enhancing annotation quality while reducing the burden on human annotators.

\subsection{Similar Example Matching}
To further augment the annotation process, we employ the principle of similar example matching. This involves leveraging previously annotated examples to inform the LLM's reasoning. Let $ D $ represent the set of previously annotated examples, and let $ E $ denote the text requiring annotation. The methodology combines the most similar examples $ E_{sim} $ with the text $ E $ that requires annotation, enabling the LLM to perform rationale-driven collaborative reasoning:
\begin{equation}
    E_{input} = (E, E_{sim})
\end{equation}

In this case, the Top-5 examples based on cosine similarity are selected from $ D $. These selected examples serve as references for the LLM during the annotation process. 
 

\section{Experiments}
\label{sec:implementation}

\subsection{Datasets}
\label{ssec:datasets}
As shown in Table \ref{tab:datasets}, the datasets utilized in this study span a range of text annotation tasks with varying levels of complexity. The Stanford Sentiment Treebank (SST-2 and SST-5) \cite{SST} provides a basis for binary and fine-grained sentiment analysis, respectively. The AG News dataset \cite{agnews} offers a straightforward multi-class annotation challenge with four distinct categories of news articles. Lastly, the DBPedia dataset \cite{dbpedia} presents a more complex annotation task with 14 classes, derived from structured information on Wikipedia. These datasets collectively enable a comprehensive evaluation of the performance of LLMs across different text annotation scenarios.

\begin{table}[]
\centering
\resizebox{86mm}{!}{
\begin{tabular}{llcc}
\hline
\textbf{Dataset} & \textbf{Task} & \textbf{$|C|$} & \textbf{Domain} \\ \hline
SST-2 & Binary sentiment annotation & 2 & Movie reviews \\
SST-5 & Fine-grained sentiment annotation & 5 & Movie reviews \\
AG News & Topic annotation & 4 & News articles \\
DBPedia & Ontology annotation & 14 & Wikipedia entries \\ \hline
\end{tabular}
}
\caption{Overview of datasets used in the experiments. $|C|$ denotes the number of classes.}
\label{tab:datasets}
\end{table}

\begin{table*}[!h]
\centering
\fontsize{9}{11}\selectfont 
\resizebox{1\textwidth}{!}{
\begin{tabular}{llcccccccc}
\hline
\textbf{Dataset} &
  \textbf{model} &
  \multicolumn{1}{l}{\textbf{ZS}} &
  \multicolumn{1}{l}{\textbf{FS}} &
  \multicolumn{1}{l}{\textbf{CoT}} &
  \multicolumn{1}{l}{\textbf{USC}} &
  \multicolumn{1}{l}{\textbf{FS-simi}} &
  \multicolumn{1}{l}{\textbf{CoT-simi}} &
  \multicolumn{1}{l}{\textbf{USC-simi}} &
  \multicolumn{1}{l}{\textbf{RDC (Ours)}} \\ \hline
\multirow{5}{*}{SST2}    & Qwen-72B         & 85.9 & 85.6          & 86.1 & 86.1          & 87.5          & \textbf{88.2} & 87.4          & \textbf{88.2}    \\
                         & Qwen-14B         & {\ul 86.7} & 85.7          & 83.7 & {\ul 86.7}    & 86.3          & 85.5          & 85.4          & \textbf{87.8} \\
                         & Qwen-7B          & 72.7 & 71.0          & 74.7 & 71.0          & {\ul 80.2}    & 73.5          & 76.1          & \textbf{85.8} \\
                         & LLaMA3-70B       & 73.1 & 72.0          & 78.8 & 78.2          & 86.9          & 78.3          & {\ul 87.1}    & \textbf{88.8} \\
                         & LLaMA3-8B        & 68.3 & 70.2          & 70.3 & 72.0          & 82.3          & 74.6          & \textbf{87.2} & {\ul 83.2}    \\ \cline{2-10} 
\textbf{}                & \textbf{Average} & 77.3 & 76.9          & 78.7 & 78.8          & {\ul 84.6}          & 80.0          & {\ul 84.6}    & \textbf{86.8} \\ \hline
\multirow{5}{*}{SST5}    & Qwen-72B         & 45.9 & 47.1          & 42.7 & {\ul 48.1}          & 46.9          & 47.1          & \textbf{48.7} & {\ul 48.1}    \\
                         & Qwen-14B         & 35.2 & 38.1          & 38.2 & \textbf{41.8} & 39.3          & 39.5          & 40.5          & {\ul 41.2}    \\
                         & Qwen-7B          & 35.8 & {\ul 37.7}          & 35.2 & 35.2          & {\ul 37.7}    & 35.8          & 37.3          & \textbf{41.4} \\
                         & LLaMA3-70B       & 47.0 & 48.9          & 46.9 & 48.4          & {\ul 51.7}    & 50.8          & 49.6          & \textbf{52.6} \\
                         & LLaMA3-8B        & 33.2 & 39.3          & 36.8 & 40.6          & \textbf{44.3} & 42.1          & 42.1          & {\ul 43.4}    \\ \cline{2-10} 
\textbf{}                & \textbf{Average} & 39.4 & 42.2          & 39.9 & 42.8          & {\ul 43.9}    & 43.1          & 43.6          & \textbf{45.3} \\ \hline
\multirow{5}{*}{AG News} & Qwen-72B         & 82.9 & 84.4          & 81.6 & 82.2          & 84.1          & 83.5          & {\ul 84.7}    & \textbf{87.4} \\
                         & Qwen-14B         & 74.4 & 76.4          & 74.6 & 74.2          & {\ul 76.4}    & \textbf{77.4} & 76.1          & 74.3          \\
                         & Qwen-7B          & 57.4 & 63.7          & 59.6 & 64.3          & {\ul 74.3}    & 72.8          & 74.1          & \textbf{76.7} \\
                         & LLaMA3-70B       & 83.2 & \textbf{86.5} & 83.8 & {\ul 84.3}    & 83.5          & 79.3          & 83.4          & 83.4          \\
                         & LLaMA3-8B        & 58.6 & 59.0          & 61.3 & 57.3          & 63.4          & {\ul 68.0}    & 65.2          & \textbf{70.8} \\ \cline{2-10} 
\textbf{}                & \textbf{Average} & 71.3 & 74.0          & 72.1 & 72.5          & 76.3          & 76.2          & {\ul 76.7}    & \textbf{78.5} \\ \hline
\multirow{5}{*}{DBPedia} & Qwen-72B         & 92.6 & \textbf{94.5} & 92.8 & 93.5          & 92.5          & {\ul 94.1}    & 91.8          & 91.8          \\
                         & Qwen-14B         & 90.5 & 91.1          & 89.8 & 91.4          & 91.5          & 91.6          & {\ul 93.6}    & \textbf{93.8} \\
                         & Qwen-7B          & 85.0 & 88.0          & 88.1 & 86.7          & 79.3          & 82.5          & {\ul 89.5}    & \textbf{90.6} \\
                         & LLaMA3-70B       & 94.1 & \textbf{95.5} & 92.5 & {\ul 95.3}    & 92.5          & 92.9          & 92.8          & 93.4          \\
                         & LLaMA3-8B        & 83.5 & 86.2          & 86.2 & 86.7          & 88.2          & 87.3          & {\ul 88.7}    & \textbf{90.1} \\ \cline{2-10} 
\textbf{}                & \textbf{Average} & 89.1 & 91.1          & 89.9 & 90.7          & 88.8          & 89.7          & {\ul 91.3}    & \textbf{91.9} \\ \hline
\end{tabular}
}
\caption{Performance comparison of different methods on various tasks. This table displays the accuracy (in \%) of different models using various methods on four benchmark datasets, with the best results in \textbf{bold} and the second-best results \underline{underlined}}.
\label{tab:performance}
\end{table*}

\subsection{Experimental Setup}
\label{ssec:experiset}
For evaluating the performance of our models across different text classification tasks, we primarily use accuracy as our metric. Meanwhile, we utilize a range of LLMs, including \textbf{Qwen1.5} (72B, 14B, 7B) and \textbf{Llama3} (70B, 8B), to comprehensively evaluate the impact of model size and architectural diversity on the effectiveness of our collaborative few-shot prompting techniques. This selection allows us to assess performance across different scales and implementations, ensuring robust and generalizable findings. 

For each model and dataset combination, we evaluate four different methods with random example: 1) Zero-Shot (ZS), 2) Few-Shot (FS) \cite{fewshotlearners}, 3) Chain-of-thought (CoT) \cite{CoT}, and 4) Universal Self-Consistency (USC) \cite{self-consistency}. Additionally, we conduct experiments using similar examples based on cosine similarity for experimental comparison: 1) Few-Shot (FS-simi), 2) Chain-of-thought (CoT-simi)  and 3) Universal Self-Consistency (USC-simi). Specifically, all experiments are run on NVIDIA A800 GPUs with VLLM inference engine. The few-shot prompting scenarios use 5 examples as references.

\subsection{Results and Analysis}
The results presented in Table \ref{tab:performance} illustrate the performance of various models across four distinct datasets, highlighting the effectiveness of our proposed methods in comparison to traditional approaches. The table categorizes the performance metrics into several annotation strategies, including Zero-Shot (ZS), Few-Shot (FS), Chain-of-Thought (CoT), Universal Self-Consistency (USC), and their respective similarity-optimized versions (FS-simi, CoT-simi, SC-simi). Notably, the results indicate that our method, referred to as RDC (Ours), consistently outperforms the baseline across nearly all models and datasets. For instance, in the SST2 dataset, the Qwen-72B model achieves a peak accuracy of 88.2\% with RDC, surpassing other methods. This trend is similarly observed in the AG News and DBPedia datasets, where the RDC method demonstrates significant improvements, particularly in the larger models. This suggests that our approach effectively leverages collaborative reasoning to enhance annotation quality, allowing for a more nuanced understanding of the data and leading to better overall performance.

Furthermore, the table highlights the variability in model performance based on the dataset and the specific task. For instance, while Qwen-14B and Qwen-7B show competitive results in certain categories, they generally lag behind the larger Qwen-72B and LLaMA3-70B models, indicating that model size and architecture play a crucial role in performance outcomes. The results also indicate that the similarity-based methods do not consistently outperform their non-similarity counterparts, suggesting that it may not enhance performance in some contexts. This variability underscores the complexity of NLP tasks, where different models may excel under different conditions. The average scores across datasets indicate that while models like Qwen-72B and LLaMA3-70B perform well, the introduction of the RDC method significantly boosts performance, particularly in challenging tasks. This finding emphasizes the importance of innovative approaches in enhancing model capabilities, as they can lead to substantial improvements in accuracy and reliability, ultimately contributing to more effective applications in real-world scenarios.

\section{Conclusion}
\label{sec:conclusion}
In this work, we present a rationale-driven LLM-based method named RDC for text annotation. RDC employs a series of iterative prompts to enable sequential collaborative data annotation across multiple rounds, demonstrating robust performance across four diverse annotation tasks by effectively leveraging the intermediate results generated by the LLM. The iterative nature of our approach mitigates the overhead associated with lengthy contexts, resulting in reduced computational expenses compared to previous studies. Although RDC requires multiple generation rounds and may incur greater overhead than traditional methods, we argue that the pursuit of high-quality annotated data is crucial for advancing AI. We are dedicated to continuously optimizing our methodology to enhance the annotation performance of LLMs, ultimately contributing to more efficient and effective data annotation processes. 
 
\bibliographystyle{IEEEbib}
\bibliography{refs}

\begin{thebibliography}{10}

\bibitem{Survey}
Li~CAI, Shu-Ting WANG, Jun-Hui LIU, and Yang-Yong ZHU,
\newblock ``Survey of data annotation,''
\newblock {\em Journal of software}, vol. 31, no. 2, pp. 302--320, 2019.

\bibitem{autoannotation}
Nicholas Pangakis, Samuel Wolken, and Neil Fasching,
\newblock ``Automated annotation with generative ai requires validation,''
\newblock in {\em ArXiv}, 2023, vol. abs/2306.00176.

\bibitem{Outperform}
Fabrizio Gilardi, Meysam Alizadeh, and Ma{\"e}l Kubli,
\newblock ``Chatgpt outperforms crowd workers for text-annotation tasks,''
\newblock {\em PNAS}, vol. 120, no. 30, pp. e2305016120, 2023.

\bibitem{chatgptannotation}
Bosheng Ding, Chengwei Qin, Linlin Liu, Lidong Bing, Shafiq Joty, and Boyang Li,
\newblock ``Is gpt-3 a good data annotator?,''
\newblock in {\em ACL (Volume 1: Long Papers)}, 2022, pp. 11173--11195.

\bibitem{raliability}
Michael~V. Reiss,
\newblock ``Testing the reliability of chatgpt for text annotation and classification: A cautionary remark,''
\newblock in {\em ArXiv}, 2023, vol. abs/2304.11085.

\bibitem{tan2024large}
Zhen Tan, Alimohammad Beigi, Song Wang, Ruocheng Guo, Amrita Bhattacharjee, Bohan Jiang, Mansooreh Karami, Jundong Li, Lu~Cheng, and Huan Liu,
\newblock ``Large language models for data annotation: A survey,''
\newblock in {\em arXiv preprint arXiv:2402.13446}, 2024.

\bibitem{hallucina}
Yue Zhang, Yafu Li, Leyang Cui, Deng Cai, et~al.,
\newblock ``Siren's song in the ai ocean: A survey on hallucination in large language models,''
\newblock in {\em ArXiv}, 2023, vol. abs/2309.01219.

\bibitem{biomedicaltask}
Israt Jahan, Md~Tahmid~Rahman Laskar, Chun Peng, and J.~Huang,
\newblock ``Evaluation of chatgpt on biomedical tasks: A zero-shot comparison with fine-tuned generative transformers,''
\newblock in {\em ACL Workshop on Biomedical Natural Language Processing}, 2023, vol. abs/2306.04504, pp. 326--336.

\bibitem{icassp1}
Jiyi Li,
\newblock ``A comparative study on annotation quality of crowdsourcing and llm via label aggregation,''
\newblock in {\em ICASSP}. IEEE, 2024, pp. 6525--6529.

\bibitem{brainstorming}
Zining Qin, Chenhao Wang, Huiling Qin, and Weijia Jia,
\newblock ``Brainstorming brings power to large language models of knowledge reasoning,''
\newblock in {\em ArXiv}, 2024.

\bibitem{self-consistency}
Xinyun Chen, Renat Aksitov, Uri Alon, Jie Ren, Kefan Xiao, Pengcheng Yin, Sushant Prakash, Charles Sutton, Xuezhi Wang, and Denny Zhou,
\newblock ``Universal self-consistency for large language models,''
\newblock in {\em ICML 2024 Workshop on In-Context Learning}.

\bibitem{dataannotationfinetune}
Keming Lu, Hongyi Yuan, Zheng Yuan, Runji Lin, Junyang Lin, Chuanqi Tan, Chang Zhou, and Jingren Zhou,
\newblock ``\# instag: Instruction tagging for analyzing supervised fine-tuning of large language models,''
\newblock in {\em ICLR}, 2023.

\bibitem{Chiang2023CanLL}
Cheng-Han Chiang and Hung yi~Lee,
\newblock ``Can large language models be an alternative to human evaluations?,''
\newblock in {\em ACL}, 2023.

\bibitem{pdfchat}
Yi~Tang, Chia-Ming Chang, and Xi~Yang,
\newblock ``Pdfchatannotator: A human-llm collaborative multi-modal data annotation tool for pdf-format catalogs,''
\newblock in {\em IUI}, 2024.

\bibitem{icassp2}
Sabyasachee Baruah and Shrikanth Narayanan,
\newblock ``Character attribute extraction from movie scripts using llms,''
\newblock in {\em ICASSP}. IEEE, 2024, pp. 8270--8275.

\bibitem{Wei2021FinetunedLM}
Jason Wei, Maarten Bosma, Vincent Zhao, Kelvin Guu, Adams~Wei Yu, Brian Lester, Nan Du, Andrew~M. Dai, and Quoc~V. Le,
\newblock ``Finetuned language models are zero-shot learners,''
\newblock in {\em ICLR}, 2022.

\bibitem{fewshotlearners}
Tom Brown, Benjamin Mann, Nick Ryder, Melanie Subbiah, Jared~D Kaplan, Prafulla Dhariwal, Arvind Neelakantan, Pranav Shyam, Girish Sastry, Amanda Askell, et~al.,
\newblock ``Language models are few-shot learners,''
\newblock in {\em NIPS}, 2020, vol.~33, pp. 1877--1901.

\bibitem{Chen2022ImprovingIF}
Mingda Chen, Jingfei Du, Ramakanth Pasunuru, Todor Mihaylov, Srini Iyer, Ves Stoyanov, and Zornitsa Kozareva,
\newblock ``Improving in-context few-shot learning via self-supervised training,''
\newblock in {\em NAACL}, 2022.

\bibitem{CoT}
Jason Wei, Xuezhi Wang, Dale Schuurmans, Maarten Bosma, Fei Xia, Ed~Chi, Quoc~V Le, Denny Zhou, et~al.,
\newblock ``Chain-of-thought prompting elicits reasoning in large language models,''
\newblock in {\em NIPS}, 2022, vol.~35, pp. 24824--24837.

\bibitem{chen2023agentverse}
Weize Chen, Yusheng Su, Jingwei Zuo, Cheng Yang, Chenfei Yuan, Chi-Min Chan, Heyang Yu, Yaxi Lu, Yi-Hsin Hung, Chen Qian, et~al.,
\newblock ``Agentverse: Facilitating multi-agent collaboration and exploring emergent behaviors,''
\newblock in {\em ICLR}, 2023.

\bibitem{SST}
Richard Socher, Alex Perelygin, Jean Wu, Jason Chuang, Christopher~D. Manning, A.~Ng, and Christopher Potts,
\newblock ``Recursive deep models for semantic compositionality over a sentiment treebank,''
\newblock in {\em EMNLP}, 2013, pp. 1631--1642.

\bibitem{agnews}
Xiang Zhang, Junbo Zhao, and Yann LeCun,
\newblock ``Character-level convolutional networks for text classification,''
\newblock in {\em NIPS}, 2015, vol.~28.

\bibitem{dbpedia}
Jens Lehmann, Robert Isele, Max Jakob, Anja Jentzsch, Dimitris Kontokostas, Pablo~N. Mendes, Sebastian Hellmann, Mohamed Morsey, Patrick van Kleef, S.~Auer, and Christian Bizer,
\newblock ``Dbpedia - a large-scale, multilingual knowledge base extracted from wikipedia,''
\newblock {\em Semantic Web}, vol. 6, pp. 167--195, 2015.

\end{thebibliography}

\end{document}